\pdfoutput=1
\documentclass[runningheads]{llncs}
\usepackage{graphicx}
\usepackage{amsmath}
\usepackage{url}
\usepackage[caption=false,font=footnotesize]{subfig}

\newcommand{\footnoteref}[1]{\textsuperscript{\ref{#1}}}

\begin{document}
\title{Facial expression and attributes recognition based on multi-task learning of lightweight neural networks}
\titlerunning{Facial expression and attributes recognition based on multi-task learning ...}
%
\author{Andrey V. Savchenko
}
\authorrunning{A.V. Savchenko}
%
\institute{HSE University, Laboratory of Algorithms and Technologies for Network Analysis, Nizhny Novgorod, Russia \\
\email{avsavchenko@hse.ru}}
\maketitle 
\begin{abstract}
In this paper, the multi-task learning of lightweight convolutional neural networks is studied for face identification and classification of facial attributes (age, gender, ethnicity) trained on cropped faces without margins. The necessity to fine-tune these networks to predict facial expressions is highlighted. Several models are presented based on MobileNet, EfficientNet and RexNet architectures. It was experimentally demonstrated that they lead to near state-of-the-art results in age, gender and race recognition on the UTKFace dataset and emotion classification on the AffectNet dataset. Moreover, it is shown that the usage of the trained models as feature extractors of facial regions in video frames leads to 4.5\% higher accuracy than the previously known state-of-the-art single models for the AFEW and the VGAF datasets from the EmotiW challenges. The models and source code are publicly available at \url{https://github.com/HSE-asavchenko/face-emotion-recognition}.

\end{abstract}
\section{Introduction}\label{sec:1}
A lot of modern intelligent systems implements facial analytics in images and videos~\cite{savchenko2018granular}, such as age, gender, ethnicity and emotion prediction~\cite{wang2021multi,mollahosseini2017affectnet}. Thousands of papers appear every year to present all the more difficult techniques and models based on deep convolutional neural network (CNN)~\cite{zhao2021computational}. Ensembles of complex CNNs won prestigious challenges and contests~\cite{bargal2016emotion,liu2020group}. 

It is known that such methods may be too complicated for their practical usage in mobile applications~\cite{demochkina2021mobileemotiface} or edge devices. As a result, there is a huge demand in development of simple easy-to-use solutions, such as multi-task learning and/or sequential training of models on different problems of facial analysis~\cite{wang2021multi,savchenko2019peerj}, that preserve the state-of-the-art quality and do not require fine-tuning on every new task and/or dataset. Unfortunately, existing techniques are rather complex due to the usage of difficult loss functions with many hyper-parameters or various tricks to distill the knowledge of previously trained neural network to pack multiple modules~\cite{hung2019increasingly}. 

The main contribution of this paper is a simplified training procedure that leads to the lightweight but very accurate CNN for multiple facial analysis tasks. The network is pre-trained on large facial dataset~\cite{cao2018vggface2}. In contrast to existing studies, it is proposed to classify carefully cropped faces using precise regions at the output of face detectors without additional margins. Though the face recognition quality becomes slightly worth when compared to training on larger facial regions, the fine-tuning of the resulted network leads to more accurate ethnicity classification and emotion recognition. Moreover, the features extracted by the latter network make it possible to achieve the state-of-the-art results among single models in video-based emotion recognition. The best trained models and Android demo application are publicly available\footnote{\url{https://github.com/HSE-asavchenko/face-emotion-recognition}}. The training code using Python 3.x with both TensorFlow 2.x and PyTorch 1.x frameworks is also available. Hence, are suitable for broad international interest and applications.

The rest part of the paper is organized as follows. A brief survey of related literature and datasets is presented in Section~\ref{sec:2}. Section~\ref{sec:3} includes detailed description of the proposed approach and training procedures. Section~\ref{sec:4} contains experimental results for facial expression classification and age, gender and ethnicity recognition. Concluding comments are discussed in Section~\ref{sec:5}.

\section{Related Works}\label{sec:2}
The best results of \textit{emotion classification} on static images are usually reported on the AffectNet dataset~\cite{mollahosseini2017affectnet}. The excellent accuracy is obtained by the pyramid with super resolution (PSR) based on large VGG-16 network~\cite{vo2020pyramid}, deep attentive center loss (DACL)~\cite{farzaneh2021facial} and the ARM method, which learns the facial representations extracted by ResNet-18 via de-albino and affinity~\cite{shi2021learning}. The state-of-the-art results for 7 emotional classes is obtained by EmotionGCN~\cite{antoniadis2021exploiting}, which exploits the dependencies between these two models using a Graph
Convolutional Network. The best accuracy for complete dataset with 8 categories is obtained by the DAN ( Distract your Attention Network)~\cite{wen2021distract} that uses feature clustering, multi-head cross attention and attention fusion networks. 

The video-based emotion recognition is typically examined on the AFEW (Acted Facial Expression In The Wild)~\cite{dhall2019emotiw} from the EmotiW (Emotion Recognition in the Wild) 2019 challenge. The DenseNet-161 was used to extract multiple facial features from each frame~\cite{liu2018multi}. Even better results are reported for an ensemble model with VGG13, VGG16 and ResNet~\cite{bargal2016emotion}. 
One of the best single models for the AFEW is obtained via the noisy student training using body language~\cite{kumar2020noisy}. Recently, the group emotions has become analyzed on the VGAF (Video-level Group AFfect)~\cite{sharma2019automatic} datasets from the EmotiW 2020. The winner of this challenge developed an ensemble of hybrid networks~\cite{liu2020group} for audio and video modalities. Remarkable performance is reached by K-injection network~\cite{wang2020implicit} and activity recognition networks~\cite{pinto2020audiovisual}.

Recognition of \textit{facial attributes}, such as age, gender, ethnicity, is typically implemented using the CNNs trained on either the IMDB-Wiki~\cite{rothe2015dex} or the UTKFace~\cite{zhang2017age}. Most of existing publicly-available age/gender prediction models, namely, MobileNet v2 (Agegendernet)\footnote{\label{Agegendernet}\url{https://github.com/dandynaufaldi/Agendernet}}, FaceNet
\footnote{\url{https://github.com/BoyuanJiang/Age-Gender-Estimate-TF}}, ResNet-50 from InsightFace ~\cite{deng2018arcface}: original and ``new" fast CNN\footnote{\url{https://github.com/deepinsight/InsightFace/}}, gender\_net and age\_net~\cite{levi2015age} trained on the Adience dataset~\cite{eidinger2014age}, 
and Deep expectation (DEX) VGG16 networks do not use large-scale face recognition datasets for pre-training. However, several papers~\cite{wang2021multi,savchenko2019peerj} clearly demonstrated the benefits of such a pre-training. The similarity among the facial processing tasks can be exploited to learn efficient face representations which boosts up their individual performances. 

There exist several studies, which use such a \textit{multi-task approach}. For instance, self-supervised co-training in multi-task learning manner reached an excellent performance in emotion recognition~\cite{pourmirzaei2021using}. Face recognition, gender identification and facial expression understanding are run simultaneously in the PAENet~\cite{hung2019increasingly} by using a continual learning approach that learns new tasks without forgetting. Unfortunately, the running-time of the best CNNs used in the above-mentioned papers is usually too high for many practical applications~\cite{savchenko2021fast}. Hence, in this paper we decided to concentrate on \textit{lightweight architectures of CNNs}, such as EfficientNet and RexNet. 

\section{Proposed Approach}\label{sec:3}
\subsection{Multi-task networks} 
\begin{figure}[t]
 \centering
\subfloat[]{
 \includegraphics[width=1.5cm,height=1.5cm]{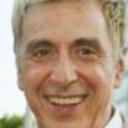}
}
\subfloat[]{
 \includegraphics[width=1.5cm,height=1.5cm]{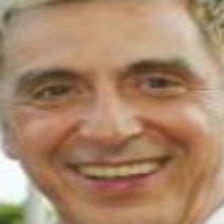}
}
\subfloat[]{
 \includegraphics[width=1.5cm,height=1.5cm]{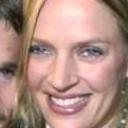}
}
\subfloat[]{
 \includegraphics[width=1.5cm,height=1.5cm]{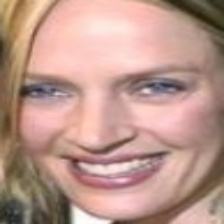}
}
 \caption{Sample images from LFW dataset: (a), (c) Center cropped; (b), (d) Cropped by MTCNN w/o margins.}
\label{fig:1}
\end{figure}

\begin{figure}[t]
\center
\includegraphics[height=8cm]{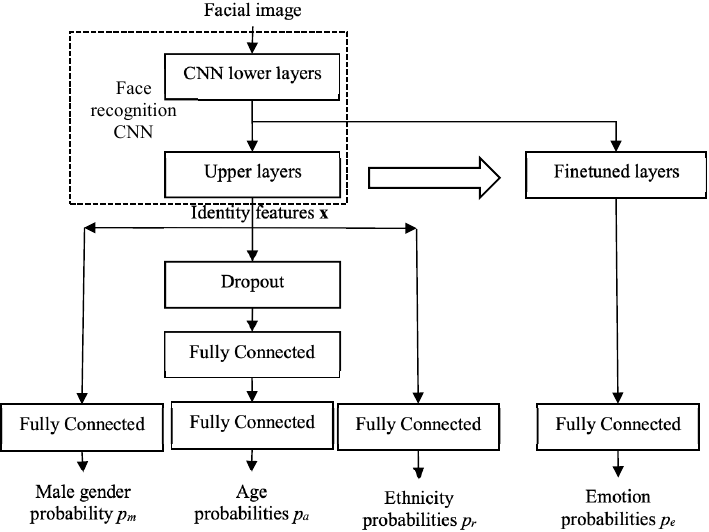}
\caption{The proposed multi-task facial expression and attributes recognition.} \label{fig:2}
\end{figure}

In this paper a multi-task neural network~\cite{savchenko2019peerj} is adapted to solve several facial attributes recognition problems (Fig.~\ref{fig:2}). The disjoint features among the tasks are exploited to increase the accuracies~\cite{wang2021multi}. At first, traditional approach is used: the base CNN is pre-trained on face identification using very large VGGFace2 dataset~\cite{cao2018vggface2}. Though the center crop of 224x224 region in each photo (Fig.~\ref{fig:1}a,c) is traditionally used as a pre-processing, it is highlighted in this paper that the higher quality is achieved using the multi-task CNN (MTCNN) face detection without (w/o) any margins (Fig.~\ref{fig:1}b,d). 

As this paper is concentrated on lightweight CNNs, it was decided to use such architectures as MobileNet, EfficientNet and RexNet as a backbone face recognition network. The resulted neural net extracts facial features $\mathbf{x}$ that are suitable to discriminate one subject from another. These features can be used to predict the attributes that are stable for a given person, i.e., gender and ethnicity, with a simple classifier, i.e., one fully connected (FC) layer. The age of the same subject is not constant but it is changed very slow. Hence, this attribute can be predicted based on the same feature vector $\mathbf{x}$, but with additional layers before the final FC layer (Fig.~\ref{fig:2}). Though the age prediction is a special case of regression problem, it is considered as a multi-class classification with $C_a$ different ages, so that it is required to predict if an observed person is 1, 2, … or $C_a$ years old~\cite{savchenko2019peerj}.

It is important to emphasize that many other facial attributes are changed rapidly, so that the facial features from face recognition should remain identical with respect to such changes. An example is the emotion recognition task: inter-class distance between face identification features of the same person with different emotions should remain much lower than the intra-class distance between different persons even with the same facial expressions. Hence, it is claimed in this paper that the facial features extracted by CNN trained on identification task cannot be directly used for emotion recognition. At the same time, lower layers of such CNN consist of feature such as edges and corners that may be better for the latter task when compared to CNN pre-trained on the dataset unrelated to faces, e.g., ImageNet. Hence, in this paper the face recognition CNN is fine-tuned on emotion dataset to either use valuable information about facial features, or predict the facial attributes that are orthogonal to the identity. 

\subsection{Training details} 
In order to simplify the training procedure, the CNNs are trained sequentially starting from face identification problem and further tuning on different facial attribute recognition tasks~\cite{hung2019increasingly}. At first, the face recognition CNN was trained using the VGGFace2 dataset. The training set contain 3,067,564 photos of 9131 subjects, while the remaining 243,722 images fill the testing set. The new head, i.e., FC layer with 9131 outputs and softmax activation, was added to the network pre-trained on ImageNet. The weights of the base net were frozen and the head was learned during 1 epoch. The categorical cross-entropy loss function was optimized using contemporary SAM (Sharpness-Aware Minimization)~\cite{foret2020sharpness} and Adam with learning rate equal to 0.001. Next, the whole CNN was trained in 10 epochs in the same way but with learning rate 0.0001. The models with the highest accuracy on validation set, namely, 92.1\%, 95.4\%/95.6\% and 96.7\% for MobileNet-v1, EfficientNet-B0/B2 and RexNet-150, respectively, were further used. 

Next, separate heads for age, gender and ethnicity prediction were added (Fig.~\ref{fig:2}) and their weights were learned. The training dataset was populated by 300K frontal cropped facial images from the IMDB-Wiki dataset~\cite{rothe2015dex} to predict age and gender~\cite{savchenko2019peerj}. Unfortunately, the age groups in this dataset are very imbalanced, so the trained models work incorrectly for faces of very young or old people. Hence, we decided to add all (15K) images from the Adience dataset~\cite{eidinger2014age}. As the latter contains only age intervals, e.g., ``(60-100)", we put all images from this interval to the average age, i.e. ``80".
The weights of the base model were frozen so that only new heads were updated. 
The binary cross-entropy loss was used for gender recognition. After 3 epochs, the resulted MobileNet-based model obtained 97\% and 13\% validation accuracies for gender and age classification, respectively. In order to make a final age prediction, only $L\in\{1,2,...,C_a\}$ indices $\{a_1, ...,a_L\}$ with the maximal posterior probabilities $p_{a_l}$ at the output of the CNN were chosen, and the mean expectation~\cite{savchenko2019peerj} is computed:

\begin{equation}
\label{eq:age_prediction}
 \overline{a} =\frac{ \sum_{l=1}^{L} {a_l \cdot p_{a_l}}}{\sum_{l=1}^{L} {p_{a_l}}}.
\end{equation}

Ethnicity classifier was trained on the subset of the UTKFace dataset~\cite{zhang2017age} with different class weights in order to achieve better performance for imbalanced classes. Conventional set of 23,708 images from the ``Aligned \& cropped faces" UTKFace was divided into 20,149 training images and 3559 testing images with $C_r=5$ races (White, Black, Asian, Indian and Latino/Middle Eastern). 

Finally, the network is fine-tuned for emotion recognition on the AffectNet dataset~\cite{mollahosseini2017affectnet}. The training set provided by the authors of this dataset contains 287,651 and 283,901 images for $C_e=8$ classes (Neutral, Happy, Sad, Surprise, Fear, Anger, Disgust, Contempt) and 7 primary expressions (the same without Contempt), respectively. The official validation set consists of 500 images per each class, i.e. 4000 and 3500 images for 8 and 7 classes. We rotate the facial images to align them based on the position of the eyes similarly to the approach from~\cite{antoniadis2021exploiting} but without data augmentation. Two ways to classify 7 emotions were studied, namely, 1) train the model on reduced training set with 7 classes; and 2) train the model on the whole training set with 8 classes, but use only 7 scores from the last (Softmax) layer. In both cases, the weighted categorical cross-entropy (softmax) loss was optimized~\cite{mollahosseini2017affectnet}:
\begin{equation}
\label{eq:emotion_loss}
 L(X,y) =-\log softmax(z_y) \cdot \underset{c \in \{1,...,C_e\}} \max N_c / N_y,
\end{equation}
where $X$ is the training image, $y \in \{1,...,C_e\}$ is its emotional class label, $N_y$ is the total number of training examples of the $y$-th class, $z_y$ is the $y$-th output of the penultimate (logits) layer, and $softmax$ is the softmax activation function. The training procedure remains similar to the initial pre-training of face recognition CNN. At first, the new head with $C_e$ outputs was added, the remaining weights were frozen and the new head was learned in 3 epochs using the SAM~\cite{foret2020sharpness}. Finally, all weights were learned during 10 epochs.

\begin{figure}[t]
 \centering
\subfloat{ \includegraphics[height=9cm]{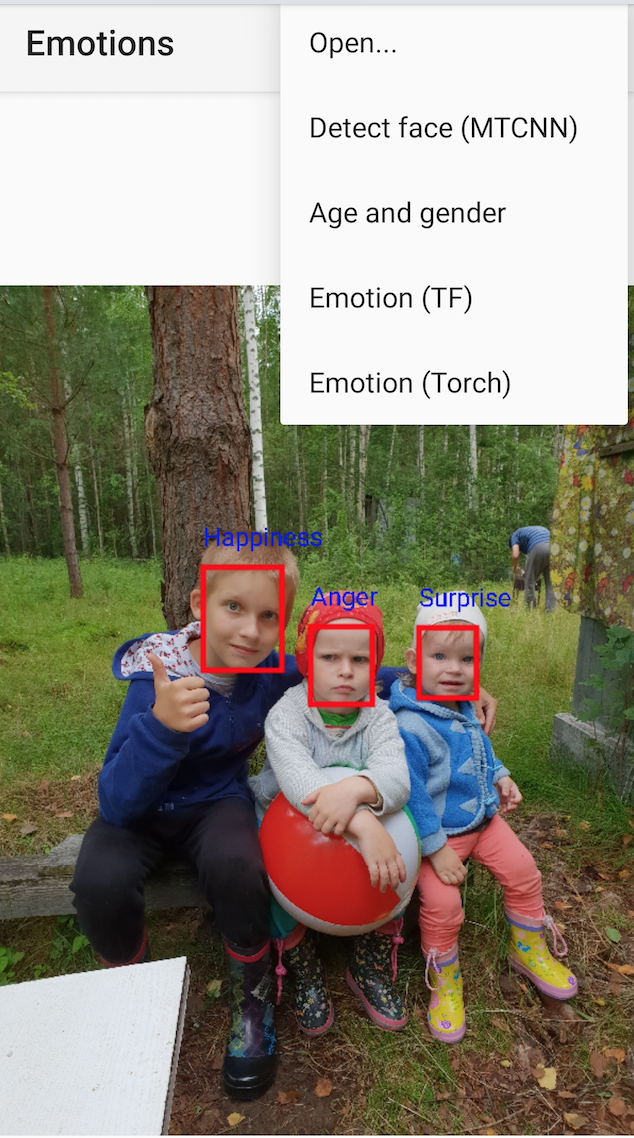}}
\subfloat{ \includegraphics[height=9cm]{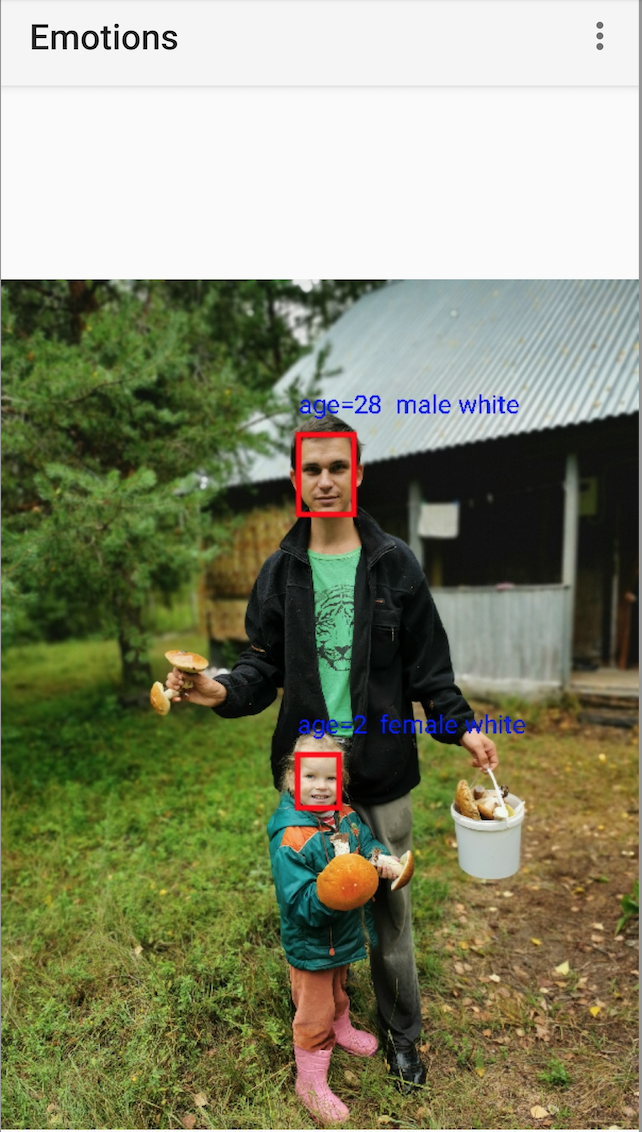}}
 \caption{Sample UI of mobile demo application.}
\label{fig:3}
\end{figure}

The obtained models are used in the mobile demo application (Fig.~\ref{fig:3}) with the publicly-available Java source code\footnote{\url{https://github.com/HSE-asavchenko/face-emotion-recognition/tree/main/mobile_app}}. It can process any photo from the gallery and predict either age/gender/ethnicity or emotional state of all detected faces.

\subsection{Video-based facial attribute recognition} 
The datasets for the video-based emotion classification tasks, i.e., the AFEW and the VGAF, contain video clips that can be used to train a classifier. The features from each frame were extracted by the networks that have been previously fine-tuned on AffectNet dataset. The largest facial region from each frame from the AFEW is fed into such CNN, and the $D$-dimensional output of its penultimate layers is stored in a frame descriptor. The video descriptor~\cite{savchenko2018granular} with dimensionality $4D$, e.g, $4 \cdot 1280=5120$ for EfficientNet-B0, is computed as a concatenation of statistical functions (mean, max, min and standard deviation)~\cite{demochkina2021mobileemotiface} applied to their frame descriptors. As a result of mistakes in face detection, facial regions have not been detected in all frames of several videos. They were completely ignored in the training set. However, though 4 validation videos does not have detected faces, they were assigned to zero descriptors with the same dimensionality as normal video descriptors, so that the validation accuracy is directly comparable with existing papers. The videos were classified with either LinearSVC or Random Forests with 1000 trees trained on the $L_2$-normed video descriptors. 

The group-level video emotion recognition task is solved similarly, though each frame may contain several facial regions. Hence, the video descriptor is computed as follows. At first, statistical functions (mean and standard deviation) of emotion features of all faces in a single frame are concatenated to obtain a descriptor of this frame. Next, all frame descriptors are aggregated using the same mean and standard deviation functions. Maximum and minimum aggregation functions are not used here in order to reduce the dimensionality of the final descriptor. Unfortunately, the clips have rather low resolution, so that only 2,619 training and 741 validation videos has at least one detected face in at least one frame. Hence, the same procedure as described for the AFEW dataset was used: the training videos without faces were ignored, and 28 validation images without faces were associated with zero descriptors.

\section{Experimental Results}\label{sec:4}
\subsection{Face identification}\label{sec:4.1}

In the first experiment, the conventional protocol of face identification~\cite{savchenko2018granular} was used for the LFW (Labeled Faces in-the-Wild) dataset. In particular, $C=596$ subjects who have at least two images in the LFW and at least one video in the YTF (YouTube Faces) database. Training set contains exactly one facial image, all other images from LFW were put into the testing set. The rank-1 face identification accuracies and their standard deviations of several CNN models pre-trained on the VGGFace2 dataset~\cite{cao2018vggface2} estimated by random 10-times repeated cross-validation are shown in Table~\ref{tab:lfw}. The models trained by the proposed pipeline (Fig.~\ref{fig:2}) are marked by italics. 

\begin{table}[t]
\caption{Rank-1 accuracy (\%) in face identification for LFW.}\label{tab:lfw}
\centering
\begin{tabular}{|c|c|c|}
\hline
CNN & Center crop & Cropped by MTCNN w/o margins\\
\hline
SENet-50~\cite{cao2018vggface2} & 97.21$\pm$ 4.19 & 96.61$\pm$ 2.02\\
\hline
\it MobileNet-v1 & 90.80$\pm$3.96 & 92.60$\pm$4.01\\
\it EfficientNet-B0 & 92.71$\pm$4.61 & 94.58$\pm$4.58\\
\it RexNet-150 & 94.76$\pm$4.45 & 96.59$\pm$3.95\\
\hline
\end{tabular}
\end{table}

As one can notice, conventional SENet-based facial descriptor~\cite{cao2018vggface2} is more accurate in this challenging task with only one training image per subject, especially if loosely cropped faces are recognized (Fig.~\ref{fig:1}a,c). However, its error rate for faces cropped by face detector without margins (Fig.~\ref{fig:1}b,d) is 0.5\% higher. In contrast to this behavior, the accuracy of the proposed descriptors is increased on 1-2\% when detected faces are classified instead of the usage of center crop. As a result, it is expected that the facial regions without background are more meaningful for other facial analytics tasks. 

\subsection{Facial Expression Recognition on Single Images}\label{sec:4.2}
In this subsection emotion classification models for 8 and 7 classes in the AffectNet dataset are studied. In addition, the models after the first training stage are analyzed, in which only the weights of the classification head were learned, while the other part of the model remains the same as in the pre-trained face recognition CNN (left part of Fig.~\ref{fig:2}). As a result, the base network extracts features appropriate for face identification. In addition, several existing models have been trained on AffectNet similarly to the models trained by proposed pipeline (Fig.~\ref{fig:2}). In particular, the following architectures have been studied: MobileNet, Inception, EfficientNet and NFNet-F0 pre-trained on ImageNet-1000 and SENet pre-trained on the VGGFace2~\cite{cao2018vggface2}. Table~\ref{tab1} gives a summary of the CNNS trained with the proposed pipeline (Fig.~\ref{fig:2}) compared with the known state-of-the-art methods. All these numbers are directly comparable with each other because they have been reported by all authors for the training and validation sets that have been determined by the authors of the AffectNet~\cite{mollahosseini2017affectnet}. The best result in each column is marked by bold.

\begin{table}[t]
\caption{Emotion recognition accuracy for AffectNet.}\label{tab1}
\centering
\begin{tabular}{|c|c|c|}
\hline
 & \multicolumn{2}{c|}{Accuracy, \% }\\
\cline{2-3}
Method & 8 classes & 7 classes\\
\hline
DAN~\cite{wen2021distract} & 62.09 & 65.69 \\
Distilled student~\cite{schoneveld2021leveraging} & 61.60 & 65.4 \\
ARM (ResNet-18)~\cite{shi2021learning} & 61.33 & 65.2 \\
PSR (VGG-16)~\cite{vo2020pyramid} & 60.68 & - \\
RAN~\cite{wang2020region} & 59.5 & - \\
Ensemble with Shared Representations~\cite{siqueira2020efficient} & 59.3 & - \\
Weighted-Loss (AlexNet)~\cite{mollahosseini2017affectnet} & 58.0 & - \\
\hline
EmotionGCN~\cite{antoniadis2021exploiting} & - & 66.34 \\
PAENet~\cite{hung2019increasingly} & - & 65.29 \\
DACL~\cite{farzaneh2021facial} & - & 65.20 \\
EmotionNet (InceptionResNet-v1)~\cite{hung2019increasingly} & - & 64.74 \\
CPG~\cite{hung2019compacting} & - & 63.57 \\
CAKE~\cite{kervadec2018cake} & - & 61.7 \\
\hline
MobileNet-v1 (ImageNet) & 56.88 & 60.4\\
Inception-v3 (ImageNet) & 59.65 & 63.1\\
EfficientNet-B0 (ImageNet) & 57.55 & 60.8\\
EfficientNet-B2 (ImageNet) & 60.28 & 64.3\\ 
NFNet-F0 (ImageNet) & 58.35 & 61.9\\
SENet-50 (VGGFace2) & 58.70 & 62.3\\
\hline
Pre-trained SENet-50 (VGGFace2) & 40.87 & 44.76\\
Pre-trained MobileNet-v1 (VGGFace2) & 40.58 & 45.17\\ 
Pre-trained EfficientNet-B0 (VGGFace2) & 49.15 & 55.87\\
Pre-trained RexNet-150 (VGGFace2) & 48.88 & 56.95\\
\hline
\it MobileNet-v1 (VGGFace2) & 60.20 & 64.71\\
\it EfficientNet-B0 (VGGFace2) & 61.32 & 64.57\\
\it EfficientNet-B0 (VGGFace2), 7 classes & - & 65.74\\
\it EfficientNet-B2 (VGGFace2) & \bf 62.42 & 66.17\\
\it EfficientNet-B2 (VGGFace2), 7 classes & - & \bf 66.34\\
\it RexNet-150 (VGGFace2), 7 classes & - & 65.54\\
\hline
\end{tabular}
\end{table}

\begin{table}[t]
\caption{Performance of emotion recognition models.}\label{tab2}
\centering
\begin{tabular}{|c|c|c|}
\hline
CNN & CPU Running time, ms & Param count, MB\\
\hline
VGG-16 & 224.7 & 134.3\\ 
ResNet-18 & 58.7 & 11.7\\ 
Inception-v3 & 160.4 & 19.7\\ 
NFNet-F0 & 621.1 & 66.4\\ 
SENet-50 & 128.4 & 25.5\\ 
\hline
MobileNet-v1 & \bf 40.6 & \bf 3.2\\
EfficientNet-B0 & 54.8 & 4.3\\ 
RexNet-150 & 104.0 & 8.3\\
\hline
\end{tabular}
\end{table}

As one can notice, the usage of a model trained on complete AffectNet training set with 8 classes for prediction of 7 emotional categories has slightly lower accuracy, though it is more universal as the same model can be used to predict either 8 or 7 emotions. Second, the experiment supports the claim that the identity features from pre-trained CNNs are not suitable for reliable facial expression recognition, though are models trained on the faces cropped by MTCNN are noticeably better. The most important property is the much higher accuracy of the models trained by the proposed approach when compared to CNNs pre-trained on ImageNet. Even the SENet model pre-trained on the VGGFace2 dataset, that has significantly higher face identification accuracy when compared to the lightweight networks (Table~\ref{tab:lfw}), is characterized by much worth emotion classification error rate. It is very important to use the face detection procedure with choice of the predicted bounding box without addition of any margins (Fig.~\ref{fig:1}). As a result, the EfficientNet-based models improved the known state-of-the-art accuracy on AffectNet for both 8 and 7 classes. 

The performance of CNNs from Table~\ref{tab1} are shown in Table~\ref{tab2}. The running time to predict emotion of one facial image was measured on the MSI GP63 8RE laptop (CPU Intel Core i7-8750H 2.2GHz, RAM 16Gb). As expected, the number of parameters and the running time of trained CNNs is also rather small, though ResNet-18 has comparable speed. 

\subsection{Video-based Facial Expression Recognition}\label{sec:4.2}
The video-based emotion classification problem is examined on two datasets from the EmotiW challenges. At first, the AFEW 8.0 dataset~\cite{dhall2019emotiw} of video clips extracted from movies is examined. The training and validation sets provided by the organizers of the EmotiW 2019 challenge contain 773 and 383 video files, respectively. Every sample belongs to one of the $C_e = 7$ emotionals (Anger, Disgust, Fear, Happiness, Sadness, Surprise, and Neutral). The facial regions in each frame were detected using the MTCNN. If it detects multiple faces in a frame, the face with the largest bounding box is selected. The validation accuracy of \textit{single} models is reported in Table~\ref{tab3}. 

\begin{table}[t]
\caption{Validation accuracy of single video-only models for AFEW.}\label{tab3}
\centering
\begin{tabular}{|c|c|}
\hline
Method & Accuracy, \%\\
\hline
Noisy student with iterative training~\cite{kumar2020noisy} & 55.17\\
Noisy student w/o iterative training~\cite{kumar2020noisy} & 52.49 \\
DenseNet-161~\cite{liu2018multi} & 51.44\\
Frame attention network (FAN)~\cite{meng2019frame} & 51.18 \\
VGG-Face~\cite{aminbeidokhti2019emotion} & 49.00\\
VGG-Face + LSTM~\cite{vielzeuf2017temporal} & 48.60\\
DSN-HoloNet~\cite{hu2017learning} & 46.47\\
LBP-TOP (baseline)~\cite{dhall2019emotiw} & 38.90\\
\hline
\it MobileNet-v1 & 55.35\\
\it EfficientNet-B0 & \bf 59.27\\
\it EfficientNet-B2 & 59.00\\
\it RexNet-150 & 57.27\\
\hline
\end{tabular}
\end{table}

As one can notice, the proposed approach provides the best known accuracy for AFEW dataset. Even the MobileNet is 0.18\% more accurate than the ResNet-18 with attention and iterative pre-training on the body language dataset. The more powerful EfficientNet-B0 architecture has 4\% higher accuracy. It practically reaches the best-known accuracy (59.42\%) of ensemble model~\cite{bargal2016emotion}, though the whole validation set is not processed due to mistakes in face detection. If only 379 validation videos with faces are analyzed, the accuracies of EfficientNet-B0 and EfficientNet-B2 are equal to 59.89\% and 59.63\%, respectively. 

Finally, group-level video-based emotion classification is studied on the recently introduced VGAF dataset~\cite{sharma2019automatic}. It has only $C_e=3$ emotion labels of a group of people, namely, Positive, Negative and Neutral. The validation set provided by the challenge's organizers contain 766 clips, while 2661 videos are available for training. The validation accuracies are presented in Table~\ref{tab4}. 

\begin{table}[t]
\caption{Validation accuracy of single video-only models for VGAF.}\label{tab4}
\centering
\begin{tabular}{|c|c|}
\hline
Method & Accuracy, \%\\
\hline
DenseNet-121 (FER+)~\cite{liu2020group} & 64.75\\
Activity Recognition Networks~\cite{pinto2020audiovisual} & 62.40\\
Inception+LSTM (baseline)~\cite{sharma2019automatic} & 52.09\\
\hline
\it MobileNet-v1 & 68.92\\ 
\it EfficientNet-B0 & 66.80\\ 
\it EfficientNet-B2 & \bf 69.84\\ 
\it RexNet-150 & 63.58\\ 
\hline
\end{tabular}
\end{table}

One can notice that the proposed approach is the best known single model for this dataset. For example, it is 2-4\% more accurate when compared to the DenseNet-121 facial model of the winner. Moreover, one can expect further improvements in group-level emotion classification by making face detection better. For example, if the models are tested on only 741 validation videos with at least one detected face, the overall accuracy is increased to 70.31\% and 68.29\% for MobileNet and EfficientNet-B0, respectively. Third, in contrast to all previous results, the accuracy of MobileNet features here is 2\% \textit{higher} when compared to EfficientNet-B0, so that both models have their advantages in various emotion recognition tasks. However, the deeper EfficientNet-B2 with higher resolution of the input image is still the best choice here. Finally, the accuracies of the trained models are higher when compared to all participants except the winner of this challenge~\cite{liu2020group} who created an ensemble of Hybrid Networks for audio and video modalities and reached excellent 74.28\% validation accuracy. For example, my model is much better than the results of the second place in this challenge~\cite{wang2020implicit}. Their ensemble of K-injection networks had accuracy 66.19\% even if audio modality was used together with the video.
 
\subsection{Facial Attributes Recognition}\label{sec:4.3}

\textit{Age/gender recognition.} In order to test the quality of age and gender prediction, the images from complete (``In the Wild") set of UTKFace~\cite{zhang2017age} were pre-processed using the following procedure from the Agegendernet\footnoteref{Agegendernet}: faces are detected and aligned with margin 0.4 using get\_face\_chip() function from DLib. Only 23,060 images with single face were used to test age and gender prediction quality. There was no fine-tuning on the UTKFace, so that the testing set contains all images from UTKFace. Results of comparison with open-source models described in Section~\ref{sec:2} are shown in Table~\ref{tab5}. 

\begin{table}[t]
\caption{Age/gender recognition results for UTKFace.}\label{tab5}
\centering
\begin{tabular}{|c|p{0.14\linewidth}|c|p{0.14\linewidth}|}
\hline
Model & Gender accuracy, \% & Age MAE & Param count, MB\\
\hline
FaceNet & 89.54 & 8.58 & 12.3\\
MobileNet v2 (Agegendernet) & 91.47 & 7.29 & 7.1\\
ResNet-50 (InsightFace) & 87.52 & 8.57 & 60.1\\
``New" model from InsightFace & 84.69 & 8.44 & 0.3\\
gender\_net & 87.32 & - & 11.2\\
VGG-16 (DEX) & 91.05 & 6.48 & 262.5 \\
\it MobileNet-v1 & 93.79 & 5.74 & 3.2\\
\hline
\end{tabular}
\end{table}

\begin{table}[t]
\caption{Ethnicity recognition accuracy (\%) for UTKFace.}\label{tab7}
\centering
\begin{tabular}{|p{0.2\linewidth}|c|c|c|c|}
\hline
Classifier& VGGFace & VGGFace-2 & FaceNet & \it MobileNet \\
\hline
Random Forest & 83.5 & 87.6 & 84.3 & 83.8 \\
SVM RBF & 78.8 & 82.4 & 86.2 & \bf 87.7 \\
LinearSVC & 79.5 & 83.1 & 85.6 & 85.6 \\
New FC layer & 80.4 & 86.4 & 84.4 & \bf 87.6 \\ 
\hline
\end{tabular}
\end{table}

The proposed model uses large-scale face recognition datasets for pre-training, so that it obtains the best results when compared to existing publicly-available models with at least 2.5\% higher accuracy of gender classification and 0.6 lower MAE (mean absolute error) of age prediction than DEX due to exploitation of the potential of very large face recognition dataset to learn face representations. Moreover, the MobileNet-based model has 80-times lower parameters when compared to two VGG-16 DEX models. 
These results are even comparable with the state-of-the-art quality for the UTKFace dataset, which is achieved by training on the part of this dataset. For instance, if the testing set with 3,287 photos of persons from the age ranges [21, 60], the MobileNet-based multi-task model achieves 97.5\% gender recognition accuracy and age prediction MAE 5.39. It is lower than 5.47 MAE of the best CORAL-CNN~\cite{cao2019consistent} on the same testing set, which was additionally trained on other subset of UTKFace. 

\textit{Ethnicity Recognition.} The UTKFace dataset was used for ethnicity recognition. The proposed MobileNet-based model was compared with traditional classification of such facial features as VGGFace (VGG-16), VGGFace-2 (ResNet-50) and FaceNet (InceptionResNet v1 trained on VGGFace-2 dataset). The validation accuracies are shown in Table~\ref{tab7}. Here the model with new head (FC layer) and SVM with RBF kernel provides an appropriate quality even in comparison with the state-of-the-art facial embeddings. 

\section{Conclusion}\label{sec:5}
In this paper the novel training pipeline (Fig.~\ref{fig:2}) was proposed that leads to the state-of-the-art accuracy of the lightweight neural networks in facial expression recognition in images and videos for several datasets. It was shown that, in contrast to existing models, additional robustness to face extraction and alignment is provided, which can be explained by pre-training of facial feature extractor for face identification from very large VGGFace2 dataset. The cropped faces with regions returned by face detectors without adding margins (Fig.~\ref{fig:1}b,d) were used. As a result, not only high accuracy (Table~\ref{tab1},~\ref{tab3},~\ref{tab4}), but also excellent speed and model size (Table~\ref{tab2}) are observed. Hence, the models trained by the proposed approach can be used even for fast decision-making in embedded systems, e.g., in mobile intelligent systems~\cite{demochkina2021mobileemotiface}. 

Though the facial representations obtained by the trained lightweight models have rather high quality, only traditional classifiers (support vector machines, random forests, etc.) have been used in this paper, so that not all of our results reach performance of the state-of-the-art methods. In future, it is necessary to improve the overall quality of facial attributes and emotion recognition by using more complex classifiers on top of extracted features, for example, by using graph convolutional networks or transformers and frame/channel-level attention ~\cite{maslov2020online}.

\section*{Acknowledgements} The work is supported by RSF (Russian Science Foundation) grant 20-71-10010.

\bibliographystyle{splncs04unsrt}
\bibliography{paper}
\end{document}